\documentclass[journal]{IEEEtran}
\usepackage{cite}
\usepackage{graphicx}

\ifCLASSINFOpdf
\else
\fi
\usepackage{amsmath}
\usepackage{algorithm}
\usepackage{algorithmic}
\usepackage{array}
\usepackage{multirow}
\usepackage{tabularx}
\newcolumntype{Y}{>{\centering\arraybackslash}X}

\begin{document}
\title{SCH-GAN: Semi-supervised Cross-modal Hashing by Generative Adversarial Network}

\author{Jian~Zhang, Yuxin~Peng and Mingkuan Yuan
\thanks{This work was supported by National Natural Science Foundation of China
	under Grants 61771025 and 61532005.}
\thanks{The authors are with the Institute of Computer Science and Technology,
	Peking University, Beijing 100871, China. Corresponding author: Yuxin Peng
	(e-mail: pengyuxin@pku.edu.cn).}}

\maketitle

\begin{abstract}
Cross-modal hashing aims to map heterogeneous multimedia data into a common Hamming space, which can realize fast and flexible retrieval across different modalities. Supervised cross-modal hashing methods have achieved considerable progress by incorporating semantic side information. However, they mainly have two limitations: (1) Heavily rely on large-scale labeled cross-modal training data which are labor intensive and hard to obtain, since multiple modalities are involved. (2) Ignore the rich information contained in the large amount of unlabeled data across different modalities, especially the margin examples from another modality that are easily to be incorrectly retrieved, which can help to model the correlations between different modalities. To address these problems, in this paper we propose a novel Semi-supervised Cross-Modal Hashing approach by Generative Adversarial Network (\textit{SCH-GAN}). We aim to take advantage of GAN's ability for modeling data distributions, so that SCH-GAN can model the distribution across different modalities, and select informative margin examples from large amount of unlabeled data to promote cross-modal hashing learning in an adversarial way. The main contributions can be summarized as follows: (1) We propose a novel generative adversarial network for cross-modal hashing. In our proposed SCH-GAN, the \textit{generative model} tries to select margin examples of one modality from unlabeled data when giving a query of another modality (e.g. giving a text query to retrieve images and vice versa). While the \textit{discriminative model} tries to distinguish the selected examples and true positive examples of the query. These two models play a minimax game so that the generative model can promote the hashing performance of discriminative model. (2) We propose a reinforcement learning based algorithm to drive the training of proposed SCH-GAN. The generative model takes the correlation score predicted by discriminative model as a reward, and tries to select the examples close to the margin to promote discriminative model by maximizing the margin between positive and negative data. Extensive experiments compared with 8 state-of-the-art methods on 3 widely-used datasets verify the effectiveness of our proposed approach.
\end{abstract}

\begin{IEEEkeywords}
Cross-modal hashing, generative adversarial network, semi-supervised.
\end{IEEEkeywords}

% \fi
\IEEEpeerreviewmaketitle

\section{Introduction}
\IEEEPARstart{W}{ith} the fast development of Internet and multimedia technologies, heterogeneous multimedia data including image, video, text and audio, has been growing very fast and enriching people's life. To make better use of such rich multimedia data, it is an important application to retrieve multimedia contents that users have interests in. Thus multimedia retrieval has attracted much attention over the past decades. However, with the explosive increase of multimedia data on the Internet, efficient retrieval from large scale databases has become an urgent need and a big challenge. To tackle this problem, many hashing methods~\cite{lsh_vldb,agh_icml,imagehashsurvey,llh_cvpr,ninh_cvpr,sdh_cvpr,sh_nips,ssh_cvpr} have been proposed to perform efficient yet effective retrieval. Generally speaking, hashing methods aim to transfer high dimensional feature into short binary codes so that similar data can have similar binary codes. Hashing methods have two major advantages when applied in multimedia retrieval: (1) Binary codes enable fast Hamming distance computation based on bit operations which can be efficiently implemented. (2) Binary codes take much less storage compared with original high dimensional feature. 

\begin{figure}[tb]
	\centering
	\includegraphics[width=0.5\textwidth]{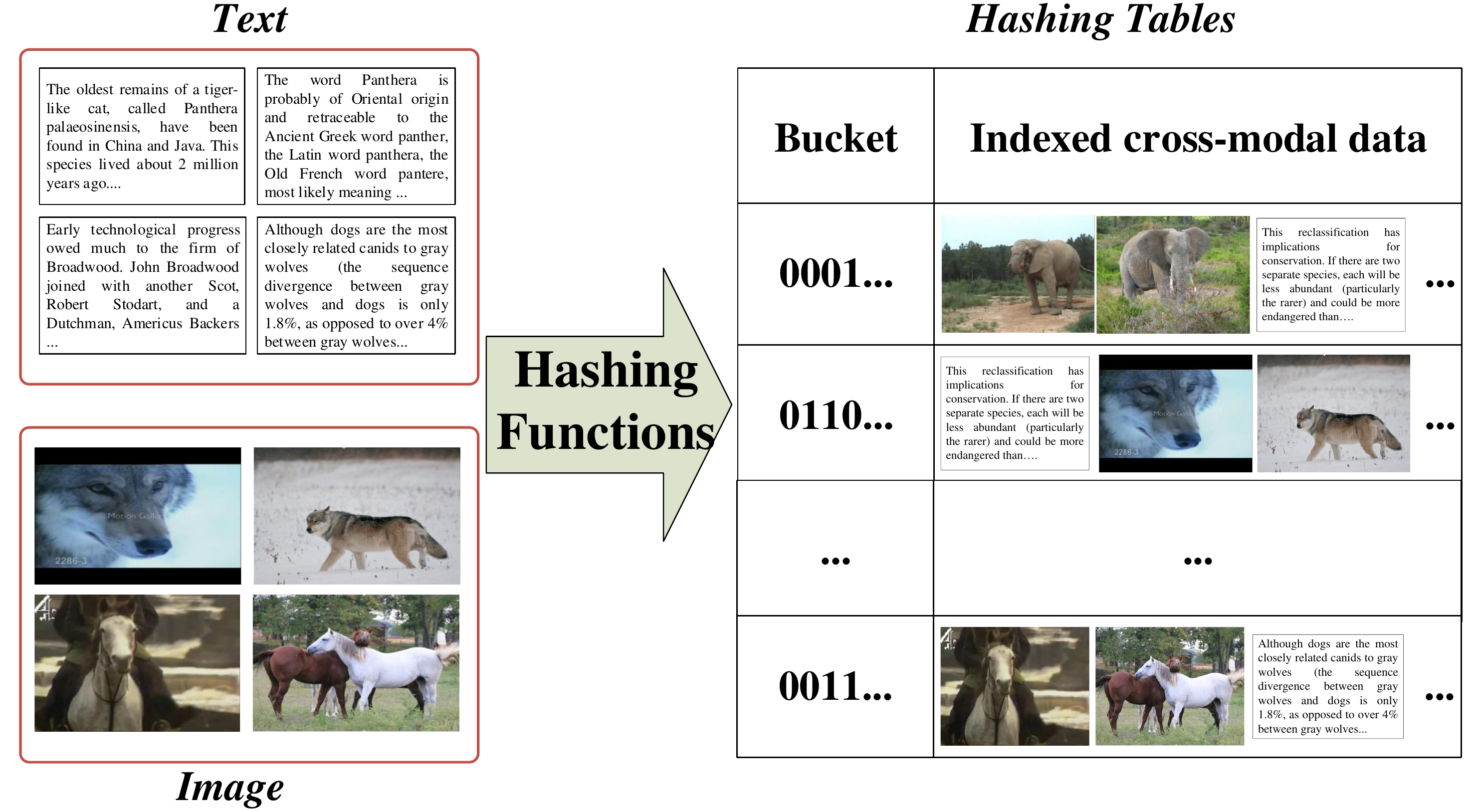}
	\caption{An example of cross-modal hashing with image and text, which projects data of different modalities into a common hamming space and performs fast retrieval.}
	\label{intro}
\end{figure}

A large amount of hashing methods are designed for single modality retrieval~\cite{lsh_vldb,agh_icml,imagehashsurvey,llh_cvpr,ninh_cvpr,sdh_cvpr,sh_nips,ssh_cvpr}, that is to say, data can only be retrieved by an query of the same modality, such as image retrieval~\cite{imagehashsurvey} and video retrieval~\cite{videohashing}. However, in real world applications, multimedia data is usually presented with different modalities. For example, an image is usually associated with a textual description, and both of them describe the same semantic. In this case, an increasing need of users is to retrieve across different modalities, such as using an image to retrieve relevant textual descriptions. Such retrieval paradigm is called cross-modal hashing. It is more useful and flexible than single modality retrieval because users can retrieve whatever they want by submitting whatever they have~\cite{crossmeidaretrieval_intro}.

The key challenge of cross-modal hashing is the ``heterogeneity gap'', which means the distribution and representation of different modalities are inconsistent, and makes it hard to directly measure the similarity between different modalities. To bridge this gap, many cross-modal hashing methods~\cite{cvh,imh,pdh,cmfh,CRH,HTH,SCM,QCH,SePH,DBRLM
,DMHOR,CMNNH,DVH,CAH} have been proposed. Generally speaking, cross-modal hashing methods can be divided into traditional methods and deep learning based methods. Traditional methods can be further divided into unsupervised methods and supervised methods based on whether semantic information is involved. The basic idea of \textit{unsupervised} cross-modal hashing methods is to project data from different modalities into a common space where correlations between them are maximized, which is similar to Canonical Correlation Analysis (CCA)~\cite{cca}. Representative unsupervised cross-modal hashing methods including Cross-view hashing (CVH)~\cite{cvh}, Inter-Media Hashing (IMH)~\cite{imh}, Predictable Dual-View Hashing (PDH)~\cite{pdh} and Collective Matrix Factorization Hashing (CMFH)~\cite{cmfh}. While \textit{supervised} cross-modal hashing methods try to learn hash functions to preserve the semantic correlations provided by labels. Representative supervised cross-modal hashing methods include Co-Regularized Hashing (CRH)~\cite{CRH}, Heterogeneous Translated Hashing (HTH)~\cite{HTH}, Semantic Correlation Maximization (SCM)~\cite{SCM}, Quantized Correlation Hashing (QCH)~\cite{QCH} and Semantics-Preserving Hashing (SePH)~\cite{SePH}. Recently, inspired by the successful applications of deep learning on image classification~\cite{imagelcass} and object recognition~\cite{deepobj}, some researches try to bridge the ``heterogeneity gap" by deep learning technique. Representative deep learning based methods include Deep and Bidirectional Representation Learning Model (DBRLM)~\cite{DBRLM}, Deep Visual-semantic Hashing (DVH)~\cite{DVH} and Cross-Media Neural Network Hashing (CMNNH)~\cite{CMNNH}.

Among the above cross-modal hashing methods, supervised methods typically achieve better retrieval accuracy due to the utilization of semantic information. However, supervised methods are very labor intensive to obtain large scale labeled training data. It is even harder to label cross-modal hashing training data, since multiple modalities are involved. Thus it is of significant importance to improve retrieval accuracy by fully exploiting unlabeled data which is very convenient to get. Traditional semi-supervised learning methods~\cite{semi} can effectively exploit distribution of unlabeled data to help supervised training. However, little efforts have been done for semi-supervised cross-modal hashing. The key challenge of semi-supervised cross-modal hashing is to exploit informative unlabeled data to promote hashing learning. With the recent progress of generative adversarial network (GAN)~\cite{gan,infogan,videogan,obgan}, which has been applied in many computer vision problems, such as image synthesis~\cite{infogan}, video prediction~\cite{videogan} and object detection~\cite{obgan}. GAN has shown its promising ability to model the data distributions. Inspired by this ability, in this paper we propose a novel semi-supervised cross-modal hashing approach by generative adversarial network (SCH-GAN). We aim to design a \textit{generative model} to fit the relevance distribution of unlabeled data near the margins between different modalities, so that it can select informative unlabeled examples close to the margin to fool the discriminative model. We also design a \textit{discriminative model} to distinguish the selected data from generative model and the true positive data, so that it can better estimate the cross-modal ranking. These two models play a minimax game to iteratively optimize each other and boost cross-modal hashing accuracy. The main contributions of this paper can be summarized as follows:
\begin{itemize}
	\item \textbf{\textit{A Generative adversarial network for cross-modal hashing (SCH-GAN)}} is proposed to fully exploit unlabeled data to improve hashing performance. In our proposed SCH-GAN, the generative model learns to fit the relevance distribution of unlabeled data, and tries to select margin examples from unlabeled data of one modality giving a query of another modality. While the discriminative model learns to judge the relevance between query and selected examples by the guidance of labeled data. Generative model tries to select the margin examples that are easily to be retrieved incorrectly to fool the discriminative model, while the latter tries to distinguish those selected examples from true positive data. Thus these two models act as two players in a minimax game, and each of them improves itself to ``beat" each other. The finally learned hashing functions from discriminative model can better reflect both semantic information of labeled data and data distributions of unlabeled data.
	\item \textbf{\textit{Reinforcement learning based loss function}} is proposed to train the generative model and discriminative model. The generative model takes the correlation score predicted by discriminative model as a reward, and tries to select the examples close to the margin to promote discriminative model to maximize the margin between positive and negative data. The discriminative model utilizes a cross-modal triplet ranking loss to learn the ranking information provided by semantic labels, and it also acts like a critic that tries to distinguish the selected examples from generative model and true positive data.
\end{itemize}

Extensive experiments on the widely-used Wikipedia, NUS-WIDE and MIRFlickr datasets demonstrate that our proposed SCH-GAN outperforms 8 state-of-the-art methods, which verify the effectiveness of SCH-GAN approach.

The rest of this paper is organized as follows. We briefly review the related works in Section II. In Section III, our proposed SCH-GAN approach is presented in detail. Then Section IV reports the experimental results as well as analyses. Finally, Section V concludes this paper.

\section{Related Works}
In this section, we briefly review some related works from two aspects: cross-modal hashing and generative adversarial network.
\subsection{Cross-modal Hashing}
Hashing methods for single modality retrieval have been extensively studied in the past decades~\cite{lsh_vldb,agh_icml,imagehashsurvey,llh_cvpr,ninh_cvpr,sdh_cvpr,sh_nips,ssh_cvpr}, and cross-modal hashing methods are receiving increasing attention in recent years. Generally speaking, most cross-modal hashing methods project data of different modalities into a common Hamming space to perform fast retrieval, and it can be divided into traditional methods and deep learning based methods. Traditional methods can be further divided into unsupervised and supervised methods based whether semantic information is involved. We will briefly review some representative works of cross-modal hashing methods.

\textit{\textbf{Unsupervised cross-modal hashing methods}} have the similar idea with Canonical Correlation Analysis (CCA)~\cite{cca}, which maps heterogeneous data into a common Hamming space to maximize the correlation. Inter-Media Hashing (IMH)~\cite{imh} is proposed to learn a common Hamming space to preserve both inter-media and intra-media consistency. Kumar et al. propose Cross-view Hashing (CVH)~\cite{cvh}, which extends image hashing method Spectral Hashing (SH)~\cite{sh_nips} to consider both intra-view and inter-view similarities with a generalized eigenvalue formulation. Rastegari et al. propose Predictable Dual-View Hashing (PDH)~\cite{pdh} for two modalities, which proposes a objective function to preserve the predictability of pre-generated binary codes, and optimize the objective function by an iterative method based on block coordinate descent. Ding et al. propose Collective Matrix Factorization Hashing (CMFH)~\cite{cmfh}, which learns unified hash codes from different modalities of one instance by collective matrix factorization with a latent factor model. Latent Semantic Sparse Hashing (LSSH)~\cite{LLSH} is proposed to use sparse coding and matrix factorization to learn separate semantic features for images and text, and then map them into a joint abstract space to reduce semantic difference. Wang et al. propose Semantic Topic Multimodal Hashing (STMH)~\cite{STMH}, which models texts as semantic topics while images as latent semantic concepts, and maps the learned semantic features for different modalities into a common semantic space, finally generates hash codes by predicting whether topics or concepts are available in the original data. Unsupervised methods try to learn cross-modal hashing functions from data distributions, thus they achieve limited accuracy on retrieving semantically similar data. To improve retrieval accuracy, some supervised methods are then proposed.
\begin{figure*}[!th]
	\centering
	\includegraphics[width=\textwidth]{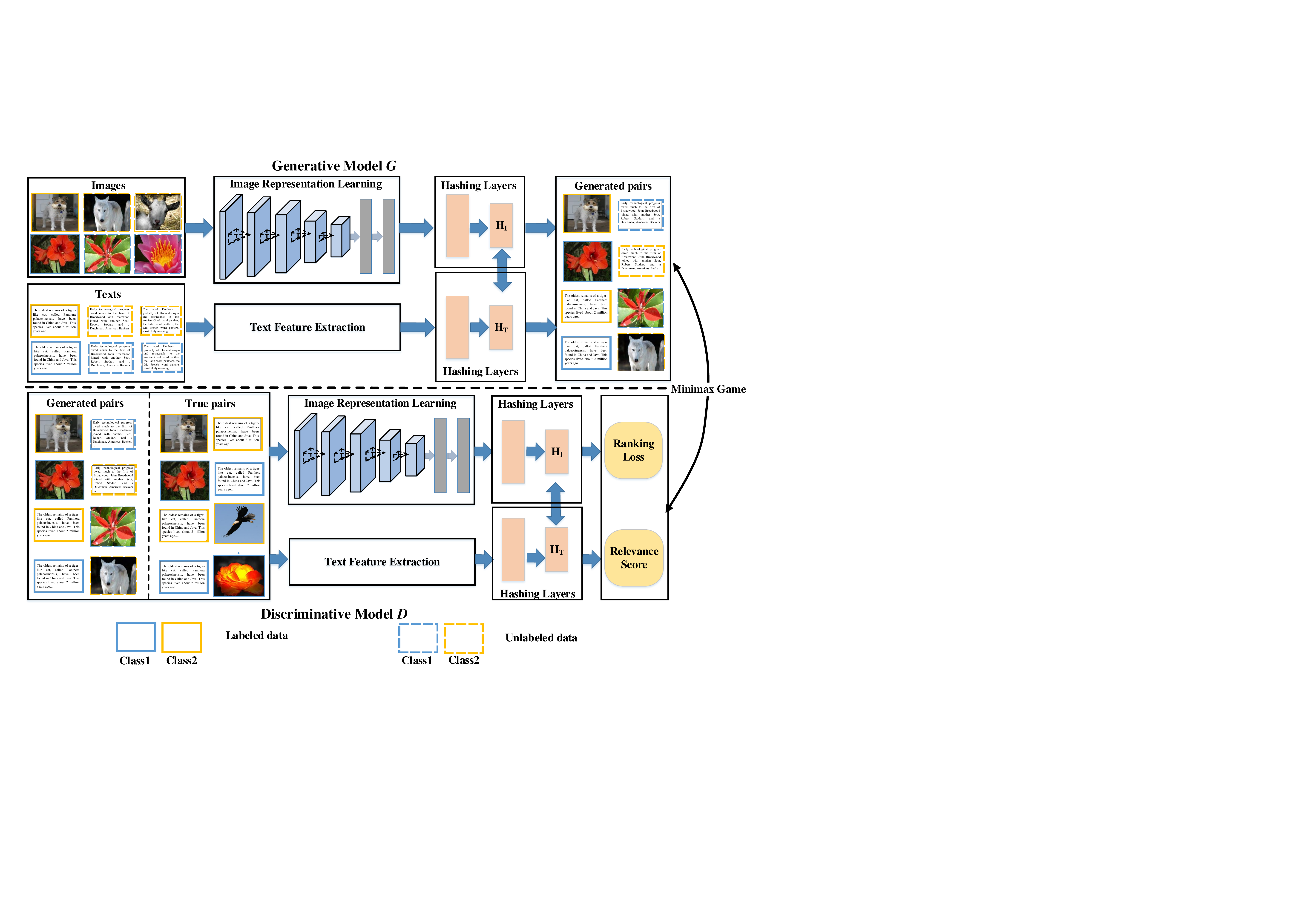}
	\caption{The overall framework of our proposed semi-supervised cross-modal hashing approach by generative adversarial network (SCH-GAN), which consists a generative model and a discriminative model. The generative model attempts to select informative unlabeled data to form a pair with a given labeled query to fool discriminative model, while the discriminative model tries to distinguish if a pair is generated or a true pair. Those two models act as two players to play a minimax game to optimize each other, and promote cross-modal hashing performance.}
	\label{framework}
\end{figure*}

\textit{\textbf{Supervised cross-modal hashing methods}} leverage semantic information obtained from labels of training data, which achieve better retrieval accuracy than unsupervised methods. Cross-modality Similarity Sensitive Hashing (CMSSH)~\cite{CMSSH} is proposed to model hashing learning as a classification problem, and can be learned in a boosting manner. Zhen et al. propose Co-Regularized Hashing (CRH)~\cite{CRH}, which learns hash functions of each bit sequentially so that the bias introduced by each hash functions can be minimized. Hu et al. propose Iterative Multi-view Hashing (IMVH)~\cite{IMVH}, which tries to learn hashing functions by preserving both within-view similarity and between-view correlations. Heterogeneous Translated Hashing (HTH)~\cite{HTH} is proposed to learn different Hamming spaces for different modalities, and then learn translators to align these spaces to perform cross-modal retrieval. Zhang et al. propose Semantic Correlation Maximization (SCM)~\cite{SCM}, which constructs semantic similarity matrix based on labels and learns hashing functions to preserve the constructed matrix. Wu et al. propose Quantized Correlation Hashing (QCH)~\cite{QCH} to simultaneously optimize cross-modal correlation and quantization error, which is also considered in many single modality hashing methods. Lin et al. propose Semantics-Preserving Hashing (SePH)~\cite{SePH}, which is a two-step supervised hashing method, it firstly transforms the given semantic matrix of training data into a probability distribution and approximates it with learned hash codes in Hamming space via minimizing the KL-divergence. Supervised cross-modal hashing methods often achieve better results than unsupervised methods because of utilization of semantic information. 

\textit{\textbf{Deep learning based methods}} are inspired by recent advance of deep neural networks, which have been applied in many computer vision problems, such as image classification~\cite{imagelcass} and object recognition~\cite{deepobj}. Zhuang et al. propose Cross-Media Neural Network Hashing (CMNNH)~\cite{CMNNH}, which learns hashing functions by preserving intra-modal discriminative capability and inter-modal pairwise correspondence. Wang et al. propose Deep Multimodal Hashing with Orthogonal Regularization (DMHOR)~\cite{DMHOR,DMHORconf}, which learns hashing functions by preserving intra-modal and inter-modal correlation, as well as reducing redundant information between hash bits. Cao et al. propose Cross Autoencoder Hashing (CAH)~\cite{CAH}, which maximizes the feature correlation of bimodal data and the semantic correlation provided by similarity labels, and CAH is based on deep autoencoder structure. Deep Visual-semantic Hashing (DVH)~\cite{DVH} is proposed, which is an end-to-end framework that integrates both feature learning and hashing function learning.

Supervised methods, especially supervised deep learning based methods have achieved promising results. However, supervised methods rely on large amount of labeled training data which are labor intensive to obtain. It is even harder to label cross-modal hashing training data, since multiple modalities are involved. Traditional semi-supervised learning methods~\cite{semi} can exploit unlabeled data effectively to help supervised training. Little efforts have been done for semi-supervised cross-modal hashing learning. In this paper, we attempt to fully exploit the unlabeled data to promote cross-modal hashing learning.

\subsection{Generative Adversarial Network}
Generative Adversarial Network (GAN)~\cite{gan} is first proposed to estimate generative model by an adversarial process. GAN consists of two models: a generative model \textit{G} that captures the data distributions, and a discriminative model \textit{D} that estimates the probability that a sample comes from real data rather than \textit{G}. These two models are trained in a adversarial way so that they compete with each other, and both of them can learn better representations of the data. Inspired by the ability of modeling data distributions of GAN, many works have attempted to apply GAN in many computer vision problems. Most popular one is image synthesis. Radford et al. propose Deep Convolutional GAN (DCGAN)~\cite{dcgan}, which adopts convolutional decoder with batch normalization and achieves better image synthesis results. Mirza et al. proposed Conditional GAN (CGAN)~\cite{cgan}, which provides side information for both generative and discriminative model to control the generated data. Inspired by CGAN, many works extend its idea to image synthesis problem, Reed et al. propose text-conditional GAN~\cite{gancls} which can generate images conditioned by textual descriptions. Odena et al. propose auxiliary classifier GAN (AC-GAN)~\cite{acgan} that generates images conditioned by class labels. Besides image synthesis, GAN is also applied to video prediction~\cite{videogan} and object detection~\cite{obgan}.

Inspired by the ability of GAN to model data distributions, in this paper we propose a novel semi-supervised cross-modal hashing by generative adversarial network (SCH-GAN). It aims design a generative model to learn the distributions of different modalities, and a discriminative model to maintain the semantic ranking information, these two models play a minimax game to iteratively optimize each other and boost cross-modal hashing accuracy.

\section{The Proposed Approach}
The overall framework of our proposed approach is demonstrated in Figure~\ref{framework}, which consists of a generative model and a discriminative model. The generative model receives the input of both labeled and unlabeled data of different modalities, and given a query of labeled data, it attempts to select informative unlabeled data around the error margins of another modality to form a pair of data. While the discriminative model receives both the generated pairs and true pairs as input, and tries to distinguish them so that it can better discriminate those margin examples. These two models are trained by playing a minimax game, and the finally trained discriminative model can be used to perform cross-modal retrieval.

\subsection{Notation}
Firstly, the formal definition of cross-modal hashing and some notations used in this paper are introduced. The two modalities involved in this paper are denoted as \textit{I} for image and \textit{T} for text. The multimodal dataset is denoted as $D=\{I,T\}, I\in \mathcal{R}^I, T\in \mathcal{R}^T$, which is further split into a retrieval database $D_{db}$ and a query set $D_{q}$. The retrieval database $D_{db}$ is also the training set, which consists of two parts, namely labeled data $D_{db}^L$ and unlabeled data $D_{db}^U = \{I_{db}^U,T_{db}^U\}$. $D_{db}^L$ is defined as $D_{db}^L = \{I_{db}^L,T_{db}^L\}$, where $I_{db}^L = \{i_p^L\}_{p=1}^n$ and $T_{db}^L = \{t_p^L\}_{p=1}^n$, $n$ is the number of labeled data. $D_{db}^L$ is also associated with corresponding class labels, which are denoted as $\{c_p^I\}_i^n$ and $\{c_p^T\}_i^n$. $D_{db}^U$ is defined as $D_{db}^U = \{I_{db}^U,T_{db}^U\}$, where $I_{db}^U = \{i_p^U\}_{p=1}^m$ and $T_{db}^U = \{t_p^U\}_{p=1}^m$, $m$ is the number of unlabeled data and $m>>n$. The query set $D_{q}$ is defined as $D_{q}=\{I_{q},T_{q}\}$, where $I_{q} = \{i_p\}_{p=1}^t$ and $T_{q} = \{t_p\}_{p=1}^t$. The goal of cross-modal hashing is to learn two mapping functions $H_I: \mathcal{R}^I\rightarrow \mathcal{R}^H$ and $H_T: \mathcal{R}^T\rightarrow \mathcal{R}^H$, so that semantically similar data of different modalities are close in the common Hamming space. Given a query of any modality, by the learned hashing functions, the semantically similar data of another modality can be retrieved by the fast Hamming distance measurement.

It is noted that supervised methods use labeled data $D_{db}^L$ to train hashing functions, however labeled data are labor intensive and hard to obtain, while unlabeled data are convenient to acquire. Thus, in this paper we attempt to further exploit the large amount unlabeled data $D_{db}^U$ to promote hashing learning. 

\subsection{Network structure}
As shown in Figure~\ref{framework}, the proposed SCH-GAN consists of a generative model and a discriminative model. We will introduce the detailed network structures of them in this subsection.

The \textit{\textbf{generative model}} is a two-pathway architecture, which receives both image and text as inputs. Each pathway consists of two parts: feature representation layers and hashing layers. For image pathway, we adopt widely-used convolutional neural networks (CNN) to represent each image, and it is noted that we keep the parameters of the CNN fixed during the training phase, since our focus is hashing function learning. While for text pathway, we use bag-of-words (BoW) features to represent textual descriptions. The structure of hash layers is the same in the two pathways, and it consists of two fully-connected layers. The first fully-connected layer serves as an intermediate layer that maps modality specific feature into a common space. The second fully-connected layer serves as hashing functions, which further map the intermediate feature into hash codes:
\begin{equation}
h(x) = sigmoid(W^Tf(x)+v)
\end{equation}
where $f(x)$ is the intermediate feature extracted from last layer, $W$ denotes the weights in the hash code learning layer, and $v$ is the bias parameter.The dimension of last fully-connected layer is set to be the same as the hash code length $q$. Through the hash code learning layer, intermediate features $f(x)$ are mapped into $[0,1]^q$. Since hash codes $h(x) \in [0,1]^q$ are continuous real values, we apply a thresholding function $g(x)=sgn(x-0.5)$ to obtain binary codes:
\begin{equation}
\label{binarycode}
b_k(x) = g(h(x)) = sgn(h_k(x)-0.5), \quad k=1,2,\cdots,q
\end{equation}
However, binary codes are hard to directly optimize, thus we relax binary codes $b(x)$ with continuous real valued hash codes $h(x)$ in the rest of this paper. Through the hashing layers, the features of different modalities are mapped into the Hamming space with same dimensions so that the similarity between different modalities can be measured by fast Hamming distance calculation. The input of generative model consists of both labeled and unlabeled data, and the goal of generative model is to select informative unlabeled data of another modality that lies around margins when given a query of one modality. This goal is achieved by the adversarial training algorithm, which will be introduced in the following subsections.

The \textit{\textbf{discriminative model}} is also a two-pathway structure, whose detailed settings are exactly the same as the generative model. The input of discriminative model is the generated (selected) relevant pairs by generative model, and the true relevant pairs sampled from labeled data. The goal of discriminative model is to distinguish whether an input pair is generated or a true pair. 

\subsection{Objective function}
Firstly, we give the formal definitions of the objectives of proposed generative model and discriminative model.
\begin{itemize}
	\item \textbf{Generative model}: $p_\theta(i^U|q_t,r)$ and $p_\theta(t^U|q_i,r)$, which try to select relevant data of one modality from unlabeled data when given a query of another modality. For example, given a text query $q_t$, the generative model tries to select relevant image $i^U$ from $I_{db}^U$. The goal of generative model is to approximate the true relevance distribution across different modalities $p_{true}(i^U|q_t,r)$.
	\item \textbf{Discriminative model}: $f_\phi(i,q_t)$ and $f_\phi(t,q_i)$, which try to predict the relevance score of the query and candidate data pair. The inputs of discriminative model consist of true pairs sampled by semantic labels, as well as generated pairs from generative model. The goal of discriminative model is to distinguish the relevant data (true pairs) and non-relevant data (generated pairs) for a query $q_i$ as accurate as possible.
\end{itemize}

Given above definitions, the generative model and discriminative model act as two players that play a minimax game: Given a query, the generative model tries to select margin data that is likely to be retrieved incorrectly to fool the discriminative model, while the discriminative model tries to distinguish between true relevant data sampled from ground-truth and the selected ones generated by its adversarial generative model. Inspired by the GAN~\cite{gan,infogan,videogan,obgan}, the adversarial process is defined as follows:
\begin{equation}
\label{minimaxgame_ti}
\begin{split}
\mathcal{V}(G,D) = &\min_{\theta}\max_{\phi}\sum_{j=1}^{n}(E_{i\sim p_{true}(i^L|q_t^j,r)}[log(D(i^L|q_t^j))]\\
&+E_{i\sim p_{\theta}(i^U|q_t^j,r)}[log(1-D(i^U|q_t^j))])
\end{split}
\end{equation}
Above equation is for text query image task, and the image query text task is similarly defined as:
\begin{equation}
\label{minimaxgame_it}
\begin{split}
\mathcal{V}(G,D) = &\min_{\theta}\max_{\phi}\sum_{j=1}^{n}(E_{t\sim p_{true}(t^L|q_i^j,r)}[log(D(t^L|q_i^j))]\\
&+E_{t\sim p_{\theta}(t^U|q_i^j,r)}[log(1-D(t^U|q_i^j))])
\end{split}
\end{equation}
These two equations are symmetric, thus in the following parts we use the text query image task as an example to present the detailed objective function. In above equations, the generative model $G$ is denoted as $p_{\theta}(i^U|q_t,r)$, which is defined as a softmax function:
\begin{equation}
\label{generativedef}
p_{\theta}(i^U|q_t,r) = \frac{\exp(-\|h_T(q_t)-h_I(i^U)\|^2)}{\sum_{i^U}\exp(-\|h_T(q_t)-h_I(i^U)\|^2)}
\end{equation}
where $h_i(\ast)$ and $h_t(\ast)$ denote the hashing functions of image pathway and text pathway respectively. Intuitively, equation~\ref{generativedef} calculates the probability that we select an image according to a given text query. It is based on the distance between image and text, smaller distance leads to larger probability. While for the true data distributions $p_{true}(i^L|q_t^j,r)$, we sample relevant images $i^L$ of query $q_t$ based on labels, such that the generative model can preserve the semantic distribution of labeled data.

The discriminative model $D$ predicts the probability of selected image $i^U$ being relevant to text query $q_t$, and $D$ is defined as the sigmoid function of relevance score:
\begin{equation}
\label{discrminativedef}
\begin{split}
D(i^U|q_t) &= sigmoid(f_\phi(i^U,q_t)) = \frac{\exp(f_\phi(i^U,q_t))}{1+\exp(f_\phi(i^U,q_t))}\\
D(i^L|q_t) &= sigmoid(f_\phi(i^L,q_t)) = \frac{\exp(f_\phi(i^L,q_t))}{1+\exp(f_\phi(i^L,q_t))}
\end{split}
\end{equation}
The relevance score of $f_\phi(i^U,q_t)$ and $f_\phi(i^L,q_t)$ are defined by triplet ranking loss as follows:
\begin{equation}
\label{triplet_u}
\begin{split}
f_\phi(i^U,q_t) = \max(0, m_i&+\|h_T(q_t)-h_I(i^+)\|^2\\
							  &-\|h_T(q_t)-h_I(i^U)\|^2)
\end{split}
\end{equation}
\begin{equation}
\label{triplet_l}
\begin{split}
f_\phi(i^L,q_t) = \max(0, m_i&+\|h_T(q_t)-h_I(i^L)\|^2\\
&-\|h_T(q_t)-h_I(i^-)\|^2)
\end{split}
\end{equation}
where $i^+$ is a semantically similar image with text query $q_t$ according to labels, $i^-$ is a semantically dissimilar image sampled from labeled data, $i^U$ is the selected image by generative model, and $m_i$ is a margin parameter which is set to be $1$ in our proposed approach. Above formulation means that we want the distance between true relevant pair $(q_t,i^+)$ smaller than that of generated pair $(q_t,i^U)$ by a margin $m_i$, so that the discriminative model can draw a clear distinguish line between the true and generated pairs. Similarly, we also want to keep the ranking based relations between labeled data.

From above objective definitions, we can observe that the generative model and discriminative model can be learned by iteratively maximizing and minimizing the same object function. The discriminative model tries to draw a margin between generated (selected) data and positive data, while the generative mode tries to select data near the margin to fool the discriminative model.
\subsection{Optimization}
By the objective function defined in equation~\ref{minimaxgame_ti}, the overall training flow of proposed approach is shown in Figure~\ref{trainignflow}. We keep the parameters of generative model fixed while training the discriminative model and vise versa. We'll introduce the optimization algorithm of these two models separately. 

\subsubsection{Optimizing discriminative model}
As shown in Figure~\ref{trainignflow}, when updating the parameters of discriminative model, we keep the generative model fixed. Firstly we use the generative model of previous iteration to select text-image and image-text pairs when given text and image as queries respectively, and we further sample true text-image and image-text pairs from labeled data. Then the discriminative model tries to \textit{maximize} the log-likelihood of correctly distinguishing the true and generated relevant pairs. When fixing generative model, based on equation~\ref{discrminativedef}, equation~\ref{minimaxgame_ti} can be rewritten as:
\begin{equation}
\label{dis_opt}
\begin{split}
\phi^\ast=&\arg\max_{\phi}\sum_{j=1}^{n}(E_{i\sim p_{true}(i^U|q_t^j,r)}[log(sigmoid(f_\phi(i^L,q_t^j)))]\\
&+E_{i\sim p_{\theta^\ast}(i^U|q_t^j,r)}[log(1-sigmoid(f_\phi(i^U,q_t^j)))])
\end{split}
\end{equation}
where $p_{\theta^\ast}$ is the generative model obtained in previous iteration. According to equations~\ref{triplet_u} and~\ref{triplet_l}, equation~\ref{dis_opt} is differentiable with respect to $\phi$, thus it can be solved by stochastic gradient descent algorithm.

\subsubsection{Optimizing generative model}
\begin{figure}[thb]
	\centering
	\includegraphics[width=0.5\textwidth]{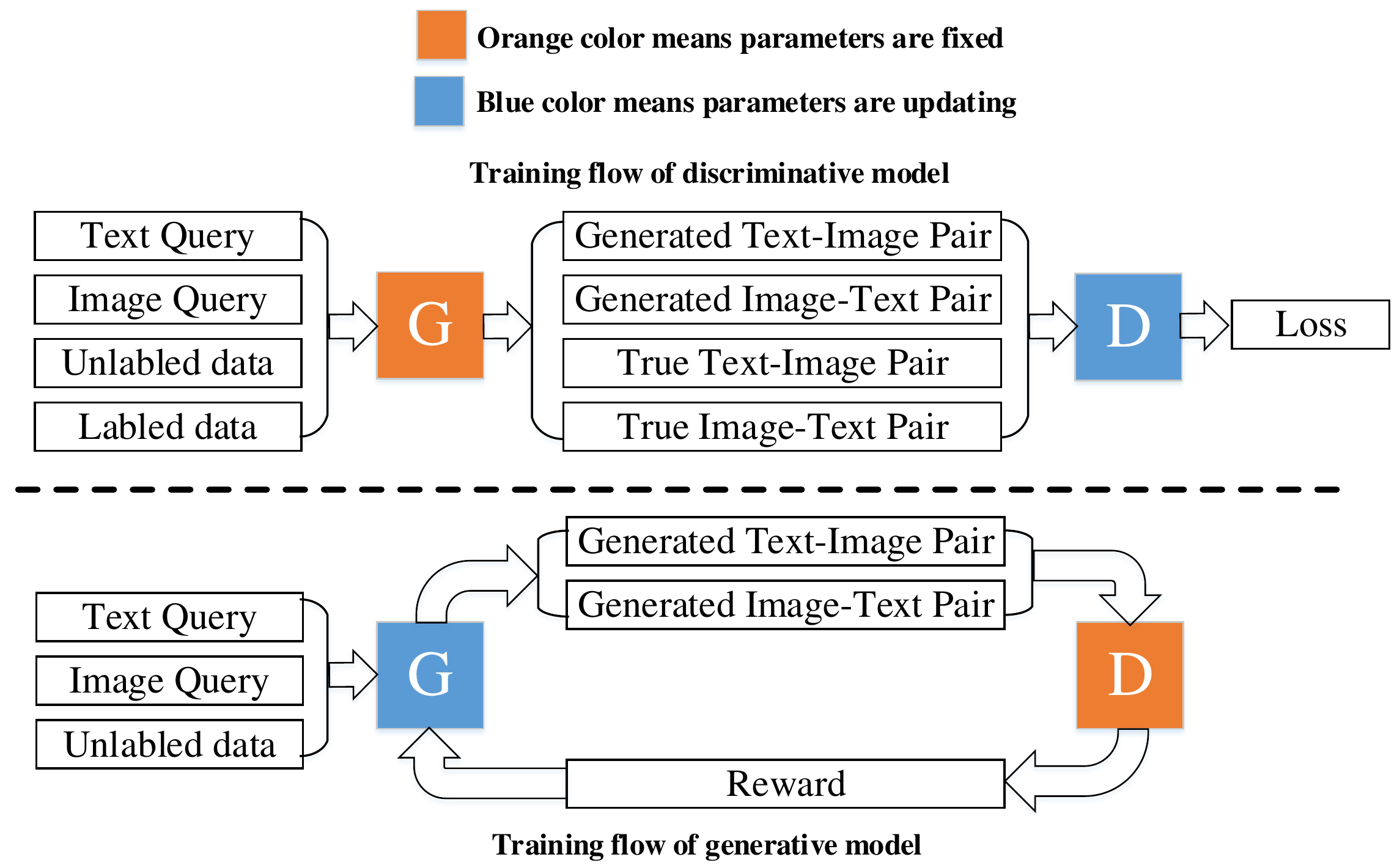}
	\caption{The training flow of the generative model and discriminative model. Best viewed in color.}
	\label{trainignflow}
\end{figure}

As demonstrated in Figure~\ref{trainignflow}, the discriminative model is fixed when training the generative model. On the contrary, the generative model tries to \textit{minimize} equation~\ref{minimaxgame_ti} and fits the true relevance distribution. When the discriminative model is fixed, the generative model can be trained by minimizing equation~\ref{minimaxgame_ti}:
\begin{equation}
\label{gen_opt}
\begin{split}
\theta^\ast=&\arg\min_{\theta}\sum_{j=1}^{n}(E_{i\sim p_{true}(i^U|q_t^j,r)}[log(sigmoid(f_{\phi^\ast}(i^L,q_t^j)))]\\
&+E_{i\sim p_{\theta}(i^U|q_t^j,r)}[log(1-sigmoid(f_{\phi^\ast}(i^U,q_t^j)))])\\
=&\arg\min_{\theta}\sum_{j=1}^{n}E_{i\sim p_{\theta}(i^U|q_t^j,r)}[log(1-\frac{\exp(f_\phi(i^U,q_t))}{1+\exp(f_\phi(i^U,q_t))})]\\
=&\arg\max_{\theta}\sum_{j=1}^{n}E_{i\sim p_{\theta}(i^U|q_t^j,r)}[log(1+\exp(f_\phi(i^U,q_t)))]
\end{split}
\end{equation}
where $f_{\phi^\ast}$ is the generative model trained in previous iteration. Different from the generative model of traditional GAN, which generates new data from continuous noise vector and can be optimized via stochastic gradient descent. The generative model of the proposed SCH-GAN selects data from unlabeled data. Since the selective strategy is discrete, it can not be optimized by stochastic gradient descent. We use policy gradient based reinforcement learning to update the parameters of generative model, which is derived as follows:
\begin{equation}
\label{gen_opt_reinforce}
\begin{split}
&\nabla_\theta E_{i\sim p_{\theta}(i^U|q_t^j,r)}[log(1+\exp(f_\phi(i^U,q_t^j)))]\\
&=\sum_{k=1}^{m}\nabla_\theta p_{\theta}(i_k^U|q_t^j,r)log(1+\exp(f_\phi(i_k^U,q_t^j)))\\
&=\sum_{k=1}^{m}p_{\theta}(i_k^U|q_t^j,r)\nabla_\theta logp_{\theta}(i_k^U|q_t^j,r)log(1+\exp(f_\phi(i_k^U,q_t^j)))\\
&=E_{i\sim p_{\theta}(i^U|q_t^j,r)}[\nabla_\theta logp_{\theta}(i^U|q_t^j,r)log(1+\exp(f_\phi(i^U,q_t^j)))]\\
&\simeq \frac{1}{m}\sum_{k=1}^{m}\nabla_\theta logp_{\theta}(i_k^U|q_t^j,r)log(1+\exp(f_\phi(i_k^U,q_t^j)))
\end{split}
\end{equation}
where $k$ denotes the $k$-th image selected by current generative model when given a text query $q_t^j$. From the perspective of deep reinforcement learning, $i_k^U$ is the action taken by policy $logp_{\theta}(i_k^U|q_t^j,r)$ according to the environment $q_t^k$, and $log(1+\exp(f_\phi(i_k^U,q_t^j)))$ acts as the reward for corresponding action, which will encourage the generative model to select data close to margins. As illustrated in Figure~\ref{trainignflow}, the rewards are calculated by the fixed discriminative network. We summarize the training process of proposed SCH-GAN in algorithm~\ref{train_alg}. It is noted that to simplify the descriptions, we take text query image task as an example in the training algorithm pseudo code.

\begin{algorithm}
	\caption{Training algorithm of proposed SCH-GAN}
	\label{train_alg}
	\begin{algorithmic}[1]
		\REQUIRE The generative model $p_{\theta}(i|q_t,r)$, the discriminative model $f_{\phi}(i,q_t)$, training data $D_{db}^L$ and $D_{db}^U$
		\STATE Randomly initialize the parameters of $p_{\theta}(i|q_t,r)$ and $f_{\phi}(i,q_t)$
		\REPEAT
		\FOR {d-step} 
		\STATE Generate $m$ text-image pairs by $p_{\theta^\ast}(i^U|q_t^j,r)$ given text query $q_t^j$
		\STATE Sampled $m$ true text-image pairs from $D_{db}^L$ based on labels
		\STATE Train discriminative model $f_{\phi}(i,q_t)$ by equation~\ref{dis_opt}
		\ENDFOR
		\FOR {g-step}
		\STATE Generate $m$ text-image pairs by $p_{\theta}(i^U|q_t^j,r)$ given text query $q_t$
		\STATE Calculate reward by $log(1+\exp(f_{\phi^\ast}(i_k^U,q_t^j)))$
		\STATE Update parameters of generative model $p_{\theta}(i^U|q_t^j,r)$ by equation~\ref{gen_opt_reinforce}
		\ENDFOR
		\UNTIL SCH-GAN converges
		\ENSURE Optimized generative model $p_{\theta^*}(i|q_t,r)$ and discriminative model $f_{\phi^*}(i,q_t)$
	\end{algorithmic}
\end{algorithm}

\subsection{Cross-modal retrieval by learned discriminative model}
It is noted that the design idea of our SCH-GAN is that the generative model tries to fit the distribution near the decision boundary, thus it is not suitable to perform cross-modal retrieval. However, the discriminative model is promoted greatly by the generative model since it can better distinguish the margin examples. Thus after the proposed SCH-GAN is trained, cross-modal retrieval can be performed by the discriminative model. More specifically, given a query of any modality (e.g. text or image), it can be first encoded into binary hash code by equation~\ref{binarycode}. Then cross-modal retrieval is performed by fast Hamming distance computation between query and each data in the database.

\section{Experiments}
In this section, we demonstrate the experimental results of our proposed SCH-GAN approach. We first introduce the datasets, evaluation metrics, implementation details and comparison methods used in our experiments. Then we compare the proposed SCH-GAN with 8 state-of-the-art methods and analyze the results. Finally, we further conduct several baseline experiments to investigate the performance of generative and discriminative model.
\subsection{Dataset}
We evaluate the proposed approach and compared methods on 3 widely-used datasets: Wikepedia~\cite{wikipedia}, NUSWIDE~\cite{nuswide} and MIRFLICKR~\cite{mirflickr}. We'll briefly introduce three datasets.
\begin{itemize}
	\item \textbf{Wikipedia} dataset~\cite{wikipedia} is a widely-used dataset for cross-modal retrieval, which is collected from ``featured articles" in Wikipedia with 10 most populated categories. This dataset consists of 2866 image/text pairs, of which the images are represented by 4096 deep features extracted from 19-layer VGGNet~\cite{vgg19}, and texts are represented by 1000 dimensional BoW (Bag of Words) features. Following~\cite{SePH}, Wikipedia dataset is split into a training set of 2173 pairs and a test set of 693 pairs. Since Wikipedia daaset is small, the training set is also used as the retrieval database, while the test set works as query.
	\item \textbf{NUSWIDE} dataset~\cite{nuswide} contains 269498 images associated with 81 concepts as ground truth, each image also has corresponding tags. Following~\cite{SePH}, we select the 10 most common concepts and the corresponding 186557 images. We take $1\%$ data of NUSWIDE dataset as the query set, and the rest as the retrieval database. As for the supervised methods, we further randomly sampled 5000 images as training set, which is similar to real world applications where only a fraction of the database are labeled. We also represent each image by 4096 deep features extracted from 19-layer VGGNet, and each texts by 1000 dimensional BoW.
	\item \textbf{MIRFlickr} dataset~\cite{mirflickr} contains 25000 images collected from Flickr, each image is also associated with textual tags and labeled with one or more of 24  provided semantic labels. Following~\cite{SePH}, we take $5\%$ of the dataset as the query set and the remaining as the retrieval database. Similar wih NUSWIDE dataset, we also randomly sample 5000 images to form the supervised training set. Similarly, we represent each image by 4096 deep features extracted from 19-layer VGGNet, and each texts by 1000 dimensional BoW.
\end{itemize}

\begin{table*}[tbh]
	\centering
	\caption{The MAP scores of two retrieval tasks on Wikipedia dataset with different length of hash codes.}
	\label{wikimap}
	\begin{tabularx}{0.85\textwidth}{c|Y|Y|Y|Y|Y|Y|Y|Y}
		\hline
		\multirow{2}{*}{Methods} & \multicolumn{4}{c|}{image$\rightarrow$text}      & \multicolumn{4}{c}{text$\rightarrow$image}      \\ \cline{2-9} 
		& 16    & 32    & 64    & 128   & 16    & 32    & 64    & 128   \\ \hline
		CVH~\cite{cvh}       & 0.193 & 0.161 & 0.144 & 0.134 & 0.297 & 0.225 & 0.187 & 0.167 \\ %\hline
		PDH~\cite{pdh}       & 0.483 & 0.483 & 0.494 & 0.497 & 0.842 & 0.842 & 0.838 & 0.851 \\ %\hline
		CMFH~\cite{cmfh}     & 0.439 & 0.496 & 0.473 & 0.461 & 0.484 & 0.548 & 0.573 & 0.568 \\ %\hline
		CCQ~\cite{CCQ}       & 0.463 & 0.471 & 0.470 & 0.456 & 0.744 & 0.788 & 0.785 & 0.741 \\ \hline
		CMSSH~\cite{CMSSH}   & 0.160 & 0.159 & 0.157 & 0.156 & 0.206 & 0.208 & 0.206 & 0.205 \\ %\hline
%		MLBE~\cite{mlbe}     &       &       &       &       &       &       &       &       \\ %\hline
		SCM\_orth~\cite{SCM} & 0.229 & 0.192 & 0.171 & 0.161 & 0.238 & 0.171 & 0.145 & 0.131 \\ %\hline
		SCM\_seq~\cite{SCM}  & 0.396 & 0.459 & 0.462 & 0.442 & 0.442 & 0.557 & 0.538 & 0.510 \\ %\hline
		SePH~\cite{SePH}     & 0.515 & 0.518 & 0.533 & 0.538 & 0.748 & 0.781 & 0.792 & 0.805 \\ \hline
%		MSAE~\cite{MSAE}     &       &       &       &       &       &       &       &       \\ %\hline
%		MDNN~\cite{MDNN}     &       &       &       &       &       &       &       &       \\ %\hline
		DCMH~\cite{DCMH}     & 0.475 & 0.508 & 0.507 & 0.503 & 0.819 & 0.828 & 0.788 & 0.720 \\ \hline
		SCH-GAN (Ours)       & 0.525 & 0.530 & 0.551 & 0.546 & 0.860 & 0.876 & 0.889 & 0.888 \\ \hline
	\end{tabularx}
\end{table*}
\begin{table*}[tbh]
	\centering
	\caption{The MAP scores of two retrieval tasks on NUSWIDE dataset with different length of hash codes.}
	\label{nusmap}
	\begin{tabularx}{0.85\textwidth}{c|Y|Y|Y|Y|Y|Y|Y|Y}
		\hline
		\multirow{2}{*}{Methods} & \multicolumn{4}{c|}{image$\rightarrow$text}      & \multicolumn{4}{c}{text$\rightarrow$image}      \\ \cline{2-9} 
		& 16    & 32    & 64    & 128   & 16    & 32    & 64    & 128   \\ \hline
		CVH~\cite{cvh}          & 0.458 & 0.432 & 0.410 & 0.392 & 0.474 & 0.445 & 0.419 & 0.398 \\ %\hline
		PDH~\cite{pdh}          & 0.475 & 0.484 & 0.480 & 0.490 & 0.489 & 0.512 & 0.507 & 0.517 \\ %\hline
		CMFH~\cite{cmfh}        & 0.517 & 0.550 & 0.547 & 0.520 & 0.439 & 0.416 & 0.377 & 0.349 \\ %\hline
		CCQ~\cite{CCQ}          & 0.504 & 0.505 & 0.506 & 0.505 & 0.499 & 0.496 & 0.492 & 0.488 \\ \hline
		CMSSH~\cite{CMSSH}      & 0.512 & 0.470 & 0.479 & 0.466 & 0.519 & 0.498 & 0.456 & 0.488 \\ %\hline
%		MLBE~\cite{mlbe}        &       &       &       &       &       &       &       &       \\ %\hline
		SCM\_orth~\cite{SCM}    & 0.389 & 0.376 & 0.368 & 0.360 & 0.388 & 0.372 & 0.360 & 0.353 \\ %\hline
		SCM\_seq~\cite{SCM}     & 0.517 & 0.514 & 0.518 & 0.518 & 0.518 & 0.510 & 0.517 & 0.518 \\ %\hline
		SePH~\cite{SePH}        & 0.701 & 0.712 & 0.719 & 0.726 & 0.642 & 0.653 & 0.657 & 0.662 \\ \hline
%		MSAE~\cite{MSAE}        &       &       &       &       &       &       &       &       \\ %\hline
%		MDNN~\cite{MDNN}        &       &       &       &       &       &       &       &       \\ %\hline
		DCMH~\cite{DCMH}        & 0.631 & 0.653 & 0.653 & 0.671 & 0.702 & 0.695 & 0.694 & 0.693 \\ \hline
		SCH-GAN (Ours)          & 0.713 & 0.726 & 0.734 & 0.748 & 0.741 & 0.743 & 0.771 & 0.779 \\ \hline
	\end{tabularx}
\end{table*}
\begin{figure*}[tbh]
	\centering
	\includegraphics[width=\textwidth]{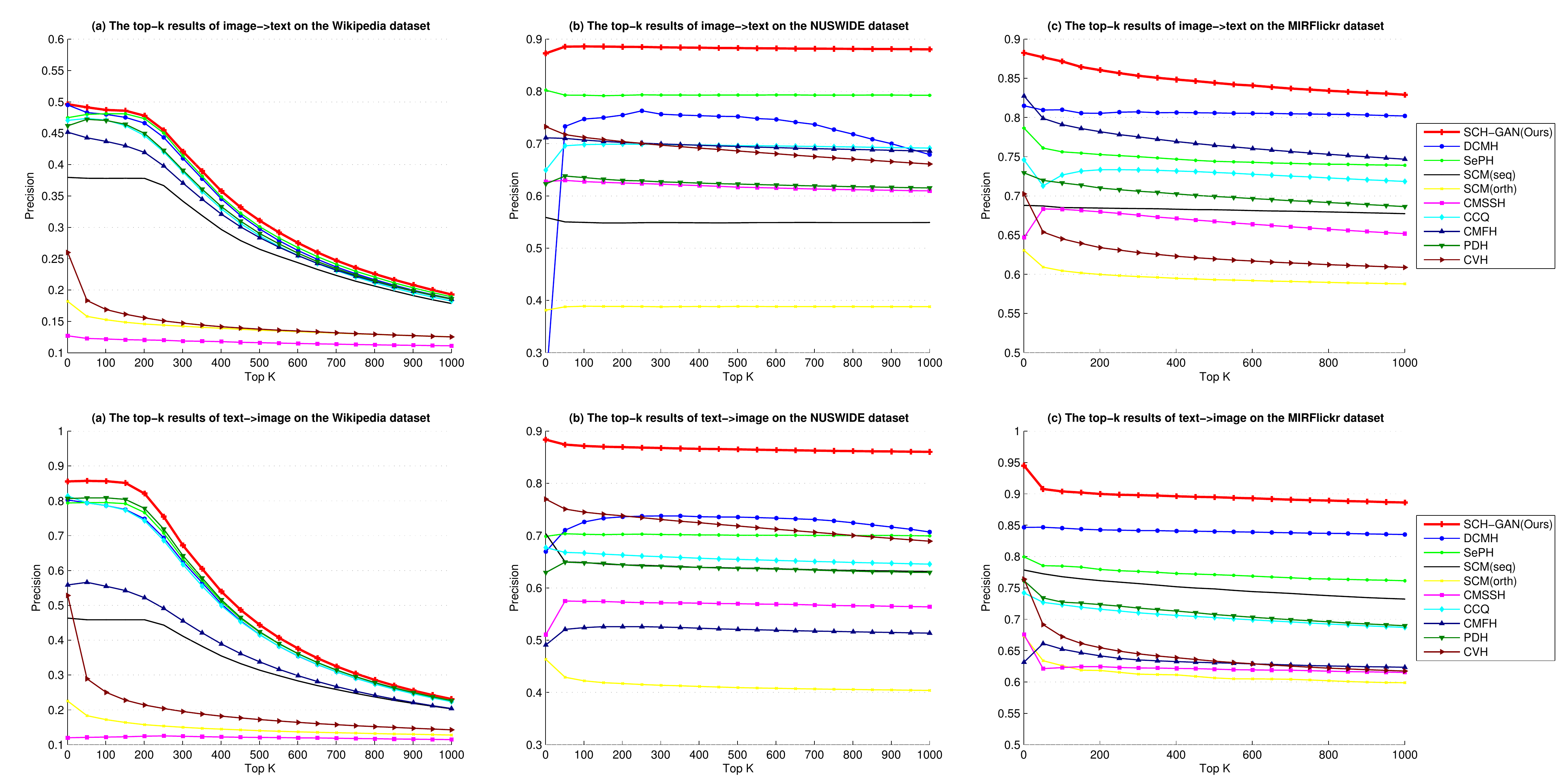}
	\caption{The top$K$-precision curves on three datasets with 64bit hash codes. The first row demonstrates the result of image query text task, while the second row shows the result of text query image task. Left, middle and right columns demonstrate Wikipedia, NUSWIDE and MIRFlickr datasets respectively.} 
	\label{topk}
\end{figure*}
\begin{table*}[tbh]
	\centering
	\caption{The MAP scores of two retrieval tasks on MIRFlickr dataset with different length of hash codes.}
	\label{mirmap}
	\begin{tabularx}{0.85\textwidth}{c|Y|Y|Y|Y|Y|Y|Y|Y}
		\hline
		\multirow{2}{*}{Methods} & \multicolumn{4}{c|}{image$\rightarrow$text}      & \multicolumn{4}{c}{text$\rightarrow$image}      \\ \cline{2-9} 
		& 16    & 32    & 64    & 128   & 16    & 32    & 64    & 128   \\ \hline
		CVH~\cite{cvh}           & 0.602 & 0.587 & 0.578 & 0.572 & 0.607 & 0.591 & 0.581 & 0.574 \\ %\hline
		PDH~\cite{pdh}           & 0.623 & 0.624 & 0.621 & 0.626 & 0.627 & 0.628 & 0.628 & 0.629 \\ %\hline
		CMFH~\cite{cmfh}         & 0.659 & 0.660 & 0.663 & 0.653 & 0.611 & 0.606 & 0.575 & 0.563 \\ %\hline
		CCQ~\cite{CCQ}           & 0.637 & 0.639 & 0.639 & 0.638 & 0.628 & 0.628 & 0.622 & 0.618 \\ \hline
		CMSSH~\cite{CMSSH}       & 0.611 & 0.602 & 0.599 & 0.591 & 0.612 & 0.604 & 0.592 & 0.585 \\ %\hline
%		MLBE~\cite{mlbe}         &       &       &       &       &       &       &       &       \\ %\hline
		SCM\_orth~\cite{SCM}     & 0.585 & 0.576 & 0.570 & 0.566 & 0.585 & 0.584 & 0.574 & 0.568 \\ %\hline
		SCM\_seq~\cite{SCM}      & 0.636 & 0.640 & 0.641 & 0.643 & 0.661 & 0.664 & 0.668 & 0.670 \\ %\hline
		SePH~\cite{SePH}         & 0.704 & 0.711 & 0.716 & 0.711 & 0.699 & 0.705 & 0.711 & 0.710 \\ \hline
%		MSAE~\cite{MSAE}         &       &       &       &       &       &       &       &       \\ %\hline
%		MDNN~\cite{MDNN}         &       &       &       &       &       &       &       &       \\ %\hline
		DCMH~\cite{DCMH}         & 0.721 & 0.729 & 0.735 & 0.731 & 0.764 & 0.771 & 0.774 & 0.760 \\ \hline
		SCH-GAN (Ours)           & 0.739 & 0.747 & 0.755 & 0.769 & 0.775 & 0.790 & 0.798 & 0.799 \\ \hline
	\end{tabularx}
\end{table*}
\subsection{Retrieval tasks and evaluation metrics}
We perform cross-modal retrieval on the 3 datasets with two kinds of retrieval tasks:
\begin{itemize}
	\item Image query text: Using image as query to retrieve semantically similar texts from retrieval database, we denote it as image$\rightarrow$text.
	\item Text query image: Using text as query to retrieve semantically similar images from retrieval database, we denote it as text$\rightarrow$image.
\end{itemize}

We utilize Hamming ranking to evaluate the proposed SCH-GAN approach and compared state-of-the-art methods. Hamming ranking gives the ranking list of a given query based on the Hamming distance, where ideal semantic neighbors are expected to be returned on the top of the ranking list. The retrieval results are evaluated based on whether the returned data and the query share the same semantic labels. We use three evaluation metrics to measure the retrieval effectiveness: Mean Average Precision (MAP), precision recall curve (PR-curve) and precision at top $k$ returned results (top$K$-precision), which are defined as follows:
\begin{itemize}
	\item Mean Average Precision (MAP): MAP is the mean value of average precisions (AP) of all queries, and AP is defined as:
	\begin{equation}
	AP=\frac{1}{R}\sum_{k=1}^{n}\frac{k}{R_k}\times rel_k
	\end{equation}
	where $n$ is the size of database, R is the number of relevant images in database, $R_k$ is the number of relevant images in the top $k$ returns, and $rel_k=1$ if the image ranked at $k$-th position is relevant and 0 otherwise. 
	\item Precision recall curve (PR-curve): The precision at certain level of recall of the retrieved ranking list, which is frequently used to measure the information retrieval performance.
	\item Precision at top $k$ returned results (top$K$-precision): The precision with respect to different numbers of retrieved samples.
\end{itemize}
Those three metrics can evaluate the proposed approach and compared methods objectively and comprehensively.

\subsection{Comparison methods}
We compare with 8 state-of-the-art methods to verify the effectiveness of our proposed approach, including unsupervised methods CVH~\cite{cvh}, PDH~\cite{pdh}, CMFH~\cite{cmfh} and CCQ~\cite{CCQ}, supervised methods CMSSH~\cite{CMSSH}, SCM~\cite{SCM} and SePH~\cite{SePH}, and deep learning based methods DCMH~\cite{DCMH}. We'll briefly introduce those compared methods.
\begin{itemize}
	\item CVH~\cite{cvh} extends Spectral Hashing (SH)~\cite{sh_nips} to considers both intra-view and inter-view similarities with a generalized eigenvalue formulation. 
	\item PDH~\cite{pdh} tries to preserve the predictability of pre-generated binary codes, and optimize the objective function by an iterative method based on block coordinate descent algorithm.
	\item CMFH~\cite{cmfh} learns unified hash codes from different modalities of one instance by collective matrix factorization with a latent factor model.
	\item CCQ~\cite{CCQ} jointly learns the correlation-maximal mappings that transform different modalities into isomorphic latent space, and learns composite quantizers that convert the isomorphic latent features	into compact binary codes.
	\item CMSSH~\cite{CMSSH} models hashing learning as a classification problem, and it is learned in a boosting manner.
	%\item MLBE~\cite{mlbe} is a probabilistic latent factor model, where the hash codes are the binary latent factors in a common Hamming space which
	%determine the generation of both intra-modality and inter-modality similarities.
	\item SCM~\cite{SCM} constructs semantic similarity matrix based on labels and learns hashing functions to preserve the constructed matrix.
	\item SePH~\cite{SePH} is a two-step supervised hashing methods, it firstly transforms the given semantic matrix of training data into a probability distribution and approximates it with learned hash codes in Hamming space via minimizing the KL-divergence.
	%\item MSAE~\cite{MSAE} utilizes stacked auto-encoder to map features of different modality into a common Hamming space by preserving the intra and inter correlations.
	%\item MDNN~\cite{MDNN} is a supervised deep learning based method, which uses deep convolutional neural network (DCNN) model and a neural language model (NLM) to learn mapping functions for the image modality and the text modality respectively.
	\item DCMH~\cite{DCMH} is an end-to-end deep learning based method, which performs feature learning and hashing function learning simultaneously.
\end{itemize}

It is noted that for a fair comparison between traditional methods and deep learning based methods, we uniformly use the same deep features as the input of traditional methods. Specifically, we use the 4096 deep features extracted from 19-layer VGGNet pre-trained on ImageNet for images, and 1000 dimensional BoW for texts. While for deep learning based methods, we use the same 19-layer VGGNet as their base network for image pathway, while keep the same settings for the text pathway.
\begin{figure*}[tbh]
	\centering
	\includegraphics[width=\textwidth]{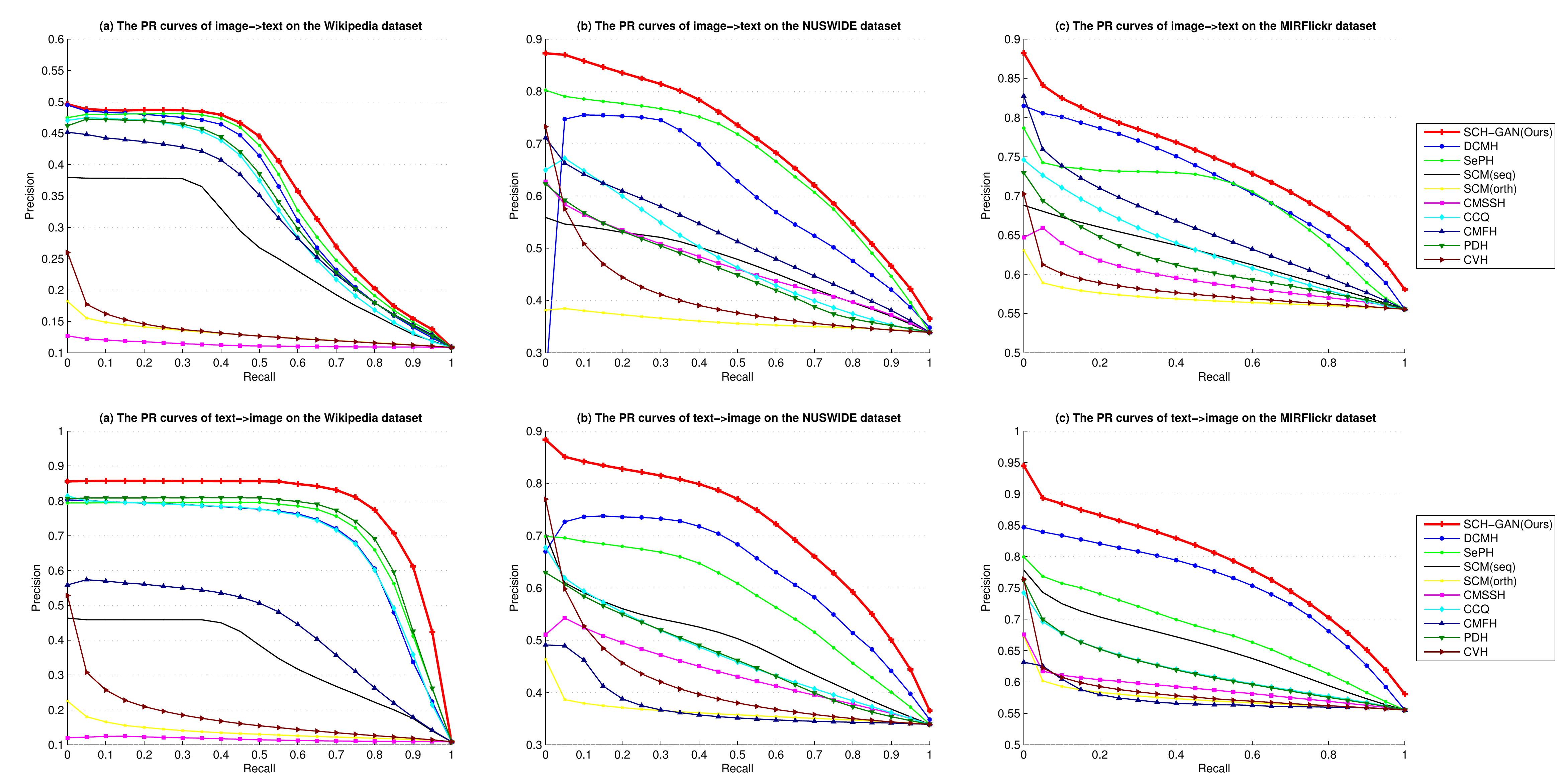}
	\caption{The precision-recall curves on three datasets with 64bit hash codes. The first row demonstrates the result of image query text task, while the second row shows the result of text query image task. Left, middle and right columns demonstrate Wikipedia, NUSWIDE and MIRFlickr datasets respectively.} 
	\label{prcurve}
\end{figure*}
\begin{table*}[]
	\centering
	\caption{Comparison between proposed approach SCH-GAN and baseline approach Dis.}
	\label{baseline}
	\begin{tabularx}{0.95\textwidth}{c|c|Y|Y|Y|Y|Y|Y|Y|Y}
		\hline
		&         & \multicolumn{4}{c|}{image$\rightarrow$text}      & \multicolumn{4}{c}{text$\rightarrow$image}      \\ \hline
		&         & 16    & 32    & 64    & 128   & 16    & 32    & 64    & 128   \\ \hline
		\multirow{2}{*}{Wikipedia} & Dis     & 0.508 & 0.494 & 0.510 & 0.510 & 0.859 & 0.847 & 0.873 & 0.858 \\ \cline{2-10} 
		& SCH-GAN & 0.525 & 0.530 & 0.551 & 0.546 & 0.860 & 0.876 & 0.889 & 0.888 \\ \hline
		\multirow{2}{*}{NUSWIDE}   & Dis     & 0.611 & 0.659 & 0.673 & 0.646 & 0.654 & 0.691 & 0.705 & 0.694 \\ \cline{2-10} 
		& SCH-GAN & 0.713 & 0.726 & 0.734 & 0.748 & 0.741 & 0.743 & 0.771 & 0.779 \\ \hline
		\multirow{2}{*}{MIRFlickr} & Dis     & 0.627 & 0.713 & 0.719 & 0.717 & 0.667 & 0.743 & 0.753 & 0.751 \\ \cline{2-10} 
		& SCH-GAN & 0.739 & 0.747 & 0.755 & 0.769 & 0.775 & 0.790 & 0.798 & 0.799 \\ \hline
	\end{tabularx}
\end{table*}
\begin{figure*}[tbh]
	\centering
	\includegraphics[width=0.9\textwidth]{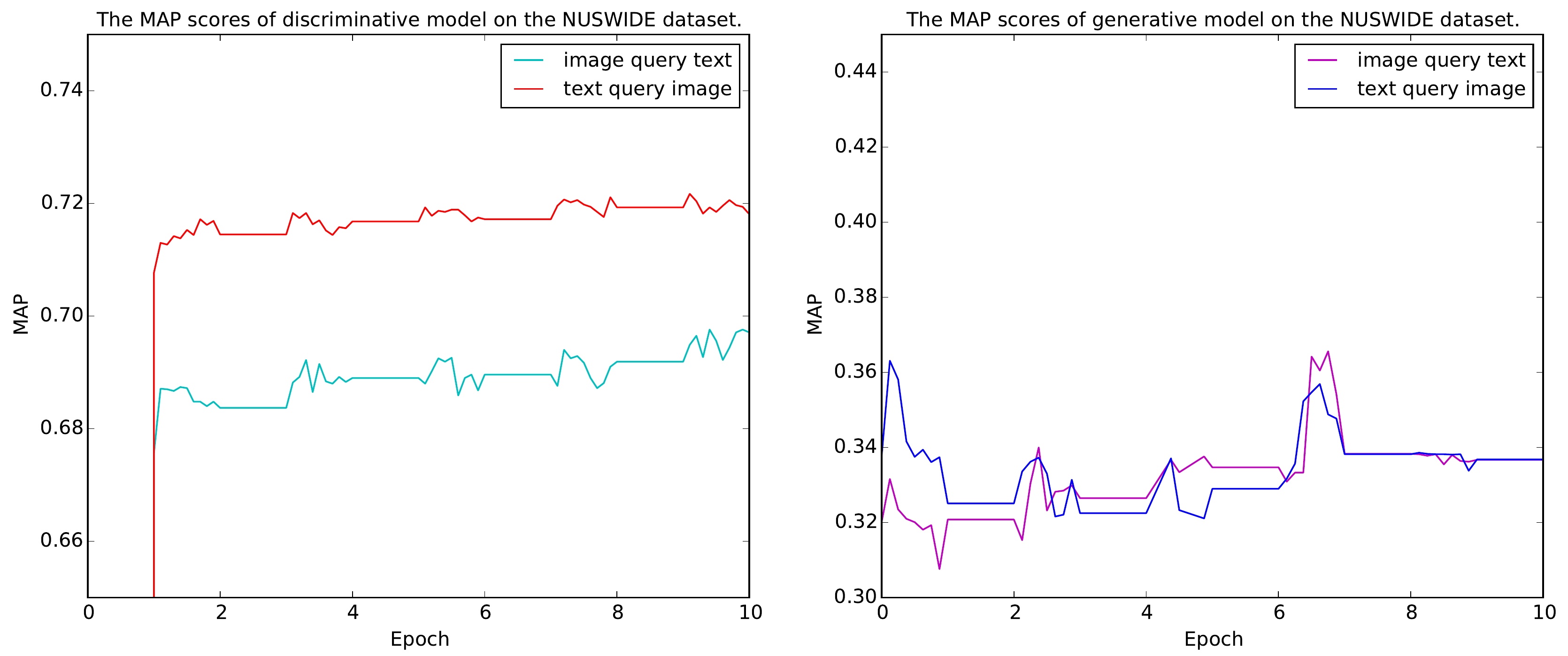}
	\caption{The retrieval accuracy of discriminative and generative model with respect to training iterations on NUSWIDE dataset with 64bit code length.} 
	\label{adversarial}
\end{figure*}
\subsection{Implementation details}
We implement the proposed SCH-GAN in Figure~\ref{framework} by tensorflow\footnote{https://www.tensorflow.org}, which is a widely-used open source software library for numerical computation using data flow graphs. The implementation details are described as follows:
\subsubsection{Data processing} For the image pathway of our proposed SCH-GAN, we use the same 19-layer VGGNet as the base network for image representation learning, and for the text pathway, we use the same 1000 dimensional BoW for text representation. It is noted that we keep the parameters of VGGNet fixed since we focus on the adversarial training of cross-modal hashing learning.
\subsubsection{Details of network} We introduce the hashing layers in Figure~\ref{framework} in detail. The hashing layers consist of an intermediate layer and a hashing layer. The intermediate layer is a fully-connected layer, whose dimension is set to be 4096 in the experiments. While the hashing layer is a fully-connect layer whose dimension is set the same as the hash code length, we also use sigmoid activation for hashing layer to force the output in the range of $[0,1]$.
\subsubsection{Training details} Here we introduce some details of the training flow demonstrated in algorithm~\ref{train_alg}. The proposed SCH-GAN is trained in a mini-batch way. The mini-batch size is set to be 64 for both image and text pathway, and $m$ in algorithm~\ref{train_alg} is set to be 20, namely we generate 20 pairs for each query. The proposed SCH-GAN is trained iteratively, specifically for each d-step and g-step, we train 1 epoch for the discriminative and generative model respectively. The learning rate of our proposed network are initialized as $0.01$, and it is decreased by a factor of 10 each two epochs.

For the compared methods, all the implementations are provided by their authors, and we follow the best settings in their papers to conduct the experiments.
\subsection{Comparison with state-of-the-art methods}
The MAP scores of two retrieval tasks on Wikepedia, NUSWIDE and MIRFlickr datasets are shown in Tables~\ref{wikimap},~\ref{nusmap} and~\ref{mirmap} respectively. From the result tables, we can observe that our proposed SCH-GAN approach achieves the best retrieval accuracy compared with state-of-the-art methods on three datasets. More specifically, the result tables are partitioned into three categories, namely unsupervised, supervised and deep learning based methods. Compared with these three categories, from the result tables we can observe that: (1) Our proposed SCH-GAN outperforms the unsupervised methods. For example, on \textit{NUSWIDE} dataset compared with best unsupervised methods CCQ~\cite{CCQ}, our proposed SCH-GAN improves the average MAP score from $0.505$ to $0.730$ on image query text task, and improves the average MAP score from $0.490$ to $0.758$ on text query image task. Similar trends can be observed on Wikipedia and MIRFlickr datasets from Tables~\ref{wikimap} and~\ref{mirmap}. It is because unsupervised methods only learn hashing functions from data distributions, which achieve limited accuracy. (2) Compared with supervised methods, our proposed SCH-GAN achieves the best results. For example, compared with best supervised method SePH~\cite{SePH} on NUSWIDE dataset, our SCH-GAN improves average MAP scores from $0.715$ to $0.730$ on image query text task, and improves from $0.654$ to $0.758$ on text query image task. Similar results can be observed on both Wikipedia and MIRFlickr datasets.  It is because our proposed SCH-GAN fully exploits the unlabeled data to promote the hash learning. (3) Our proposed SCH-GAN also outperforms deep learning based methods, which improves average MAP scores from $0.652$ (DCMH~\cite{DCMH}) to $0.730$ on image query text task, and from $0.696$ to $0.758$ on text query image task. It demonstrates that the generative model can  promote the discriminative model by challenging it with hard examples around margins.

Figures~\ref{topk} and~\ref{prcurve} show the top$K$-precision and precision-recall curves on the three datasets with 64bit code length. We can observe that on both image query text and text query image tasks, our proposed SCH-GAN achieves the best accuracy, which further demonstrates the effectiveness of our proposed approach.

\subsection{Baseline experiments}
We conduct two baseline experiments to demonstrate the performance of generative and discriminative model. Firstly, we investigate the retrieval performance of generative and discriminative model to give more insight of adversarial training. Secondly, we compare proposed SCH-GAN with a baseline approach without adversarial training to verify its effectiveness.
\subsubsection{Performance of adversarial training} We demonstrate the retrieval accuracy of generative and discriminative models during the training process. The result is shown in Figure~\ref{adversarial}, we can observe that during the adversarial training, the accuracy of discriminative model is increasing after the generative model updated. It means that the generative model selects more informative examples for the discriminative model to promote its accuracy.
\subsubsection{Comparison with baseline approach} In our proposed approach, the discriminative model can be trained solely without generative model by using the triplet ranking loss in equation~\ref{triplet_l}. This is considered as a simple supervised baseline approach without using adversarial training, we denote this baseline approach as $Dis$. Compare SCH-GAN with Dis, we can verify the effectiveness of adversarial training. The results are shown in Table~\ref{baseline}, and we can observe that the proposed SCH-GAN constantly outperforms baseline method $Dis$ on all the three datasets. It demonstrates that the generative model selects informative margin examples to promote the training of discriminative model, thus promotes the accuracy of cross-modal hashing.

\section{Conclusion}
In this paper we have proposed a novel semi-supervised cross-modal hashing approach based generative adversarial network (SCH-GAN). Firstly, we propose a novel generative adversarial network to model cross-modal hashing. In our proposed SCH-GAN, the \textit{generative model} tries to select margin examples of another modality from unlabeled data when giving a query of one modality (e.g. giving a text query to retrieve images and vice versa). While the \textit{discriminative model} tries to predict the correlation between query and selected examples of generative model. These two models play a minimax game to iteratively optimize each other in an adversarial way.
Secondly, we propose a reinforcement learning based algorithm to drive the training of proposed SCH-GAN. The generative model takes the correlation score predicted by discriminative model as a reward, and tries to select the examples close to the margin to promote discriminative model by maximizing the margin between positive and negative data. Experiments compared with 8 state-of-the-art methods on 3 widely-used datasets verify the effectiveness of our proposed approach.

In the future works, on one hand, we attempt to extend current framework to an unsupervised scenario which is more generalized. On the other hand, we will extend current approach to other modalities such as video and audio so that it can exploit complex correlations between multiple modalities.

\ifCLASSOPTIONcaptionsoff
  \newpage
\fi



\bibliographystyle{IEEEtran}
\bibliography{CMHGAN.bib}
\end{document}